%% file: main.tex
\documentclass[10pt,twocolumn,letterpaper]{article}

\usepackage{cvpr}
\usepackage{times}
\usepackage{epsfig}
\usepackage{graphicx}
\usepackage{amsmath}
\usepackage{amssymb}
\usepackage{booktabs} % For formal tables
\usepackage{subcaption}
\usepackage{xurl}
\usepackage{array}
\usepackage{tabu}
\usepackage{multirow}
\usepackage{array}
\usepackage{algorithm2e}
\usepackage{enumitem}
\usepackage{textcomp}
\graphicspath{ {./images/} }

\cvprfinalcopy % *** Uncomment this line for the final submission

\begin{document}

%%%%%%%%% TITLE
\title{Evaluating Deep Learning in SystemML using Layer-wise Adaptive Rate Scaling (LARS) Optimizer}

\author{Kanchan Chowdhury\\
Arizona State University\\
Tempe, Arizona\\
{\tt\small kchowdh1@asu.edu}
\and
Ankita Sharma\\
Arizona State University\\
Tempe, Arizona\\
{\tt\small ashar236@asu.edu}
\and
Arun Deepak Chandrasekar\\
Arizona State University\\
Tempe, Arizona\\
{\tt\small achand66@asu.edu}
}

\maketitle
%\thispagestyle{empty}

\input{abstract}

\input{introduction}

\input{related_work}

\input{methodology}

\input{evaluation}

\input{conclusion}

\input{future-work-challenges}

{\small
\bibliographystyle{ieee_fullname}
\bibliography{main}
}

\end{document}

%% file: abstract.tex
%%%%%%%%% ABSTRACT
\begin{abstract}
Increasing the batch size of a deep learning model is a challenging task. Although it might help in utilizing full available system memory during training phase of a model, it results in significant loss of test accuracy most often. LARS\cite{lars} solved this issue by introducing an adaptive learning rate for each layer of a deep learning model. However, there are doubts on how popular distributed machine learning systems such as SystemML\cite{systemml} or MLlib\cite{mllib} will perform with this optimizer. In this work, we apply LARS optimizer to a deep learning model implemented using SystemML.We perform experiments with various batch sizes and compare the performance of LARS optimizer with \textit{Stochastic Gradient Descent}. Our experimental results show that LARS optimizer performs significantly better than Stochastic Gradient Descent for large batch sizes even with the distributed machine learning framework, SystemML.
\end{abstract}

%% file: introduction.tex
%%%%%%%%% BODY TEXT
\section{Introduction}
Applications of deep learning to solve problems such as classification, regression and prediction have seen tremendous growth in this era because of its success in solving these problems. The availability of large-scale datasets and high-performance computing devices have also enlarged the scope of deep learning algorithms. Although deep learning is helping us find automated solutions to many real-world problems, the time required for training a large deep learning model has always been a bottleneck. It is seen that some deep learning models even take several days to finish. The time required for training a deep learning model depends mainly on two things: volume of the dataset and size of the model.

Nowadays, because of the availability of various sensor devices, data volume is increasing rapidly. It is a common and also required practice to feed a deep learning model with a huge volume of dataset. This immense size of training data results in long training time which is a bottleneck of deep learning models. Besides the size of the dataset, various giant technology companies need to train deep learning models with a lot of layers and parameters. For example, a popular deep learning model, ResNet-50\cite{resnet} has around 25.6 million parameters. This numerous number of parameters is another reason for slow deep learning training.

As data size increases, deep learning models need to process a large set of data in memory. Today, because of the improvement of computing devices, feeding large data into memory is not a problem. Google and some other companies have computing devices that are so powerful that even the largest deep learning models cannot make full use of these computing devices. Considering that these super-powerful computing devices are available, is it possible to reduce the training time of the largest deep learning models significantly without affecting the accuracy?

It is observed that an increase in the batch size results in a decrease in total training time. Increasing the batch size up to a certain threshold is also good for the accuracy of deep learning. Another important fact to note is that increasing the batch size indefinitely reduces the model accuracy significantly which prevents deep learning practitioners to keep the batch size within reasonable limit although they might have powerful computing devices to feed the data into memory. Besides increasing the batch size, another possible solution to reduce the training time is parallelizing deep learning. Training process of a deep learning model can be parallelized in two ways: parallelizing data and parallelizing models. Machine learning systems such as TensorFlow allow us to parallelize deep learning training. Besides, we also have distributed systems such as Apache Spark\cite{apache-spark}. Some other machine learning libraries are also available on top of Apache Spark which include but are not limited to SystemML\cite{systemml}, MLlib\cite{mllib}, etc. All these distributed systems are helping us parallelize deep learning training to reduce training time. Is it possible to further reduce the training time as well as increase the batch size while maintaining the test accuracy of the model?

As we discussed earlier, training of a deep learning model can be parallelized through either data parallelism or model parallelism. We cannot only rely on data parallelism because the model size (number of parameters and layers) is too big sometimes which makes model parallelism a mandatory option. Again, model parallelism can be done in two ways: 1) parallelize within each layer and 2) parallelize across different layers. Each of these approaches has problems. The problem with parallelizing within each layer is that the wide model is not efficient. A deep model can be better than a wide model. Also, the problem with parallelizing across different layers is that parallel efficiency is low which is 1/P. So, the solution is combining model parallelism with data parallelism.

Scaling a machine learning model is a difficult task. Most often, it results in the generalization problem. The generalization problem indicates we might end up in high training accuracy while test accuracy is very low. A promising solution can be auto-tuning the learning rate while scaling the batch size. Another idea is using different learning rates for different layers. This idea of layer-wise adapting the learning rate for increased batch size was first introduced by LARS\cite{lars} for deep learning in systems such as TensorFlow. It showed that batch size can be increased significantly even on the CPU by the use of a layer-wise adaptive learning rate technique.

A standard value of learning rate is effective to train a model perfectly, but it always results in a long training time. Again, a larger value learning rate speeds up the training process, but it might result in under-fitting the model. Learning rate decay is a proven solution to mitigate this problem. The idea of learning rate decay is to start with an initial learning rate and to change the learning rate after every epoch at a constant rate. The problem is updating the learning rate at a constant rate after every epoch does not help much which inspires the invention of algorithms such as LARS\cite{lars} and LAMB\cite{lamb}. Inspired by the success of LARS and LAMB, our proposal is to apply these algorithms to speed up machine learning on top distributed machine learning systems such as SystemML. For this phase of our project, we focus only on LARS optimizer instead of trying both LARS and LAMB optimizers.

We selected SystemML in this project for two reasons. Firstly, this distributed machine system is not yet as optimized as TensorFlow or PyTorch for deep learning. The LARS implementation is not available for this framework. Our second reason for selecting SystemML is that it can be used on top of Apache Spark. Although distributed deep learning is supported by TensorFlow and PyTorch, Apache Spark seems to be very effective for inter-cluster communication. This is because Apache Spark is mainly developed for distributed and large scale systems. Spark RDDs are very fast and effective for remote memory access. The summary of our contribution is outlined below:
\begin{itemize}[noitemsep]
\item Applying LARS optimizer to a deep learning model implemented using SystemML.
\item Comparing test accuracy of LARS optimizer with that of Stochastic Gradient Descent.
\item Evaluating test and training accuracy by increasing batch size.
\item Reducing the generalization error using LARS optimizer.
\item Finding the challenges in applying LARS optimization technique to SystemML.
\end{itemize}

%% file: related_work.tex
\section{Related Work}
The use of machine learning and deep learning-based algorithms has been increasing for a long time for various applications. Various platforms like Tensorflow, PyTorch, and Keras provide extensive functionality in order to support data parallelism and model parallelism. While data parallelism has its own disadvantage of not getting fit if the model is too large even if it can support accelerating the training process. The model parallelism has been proposed in various applications in order to solve this issue. Recently model parallelism has been successfully implemented for relational databases\cite{declarative-rdbms}. This computation on an RDBMS has been implemented by applying machine learning-based algorithms. They proposed that different parts of the model can be stored in a set of tables and SQL queries can be used for computation purposes.  RDBMS has been studied widely for distributed computing for a long time. The paper has also discussed challenges in involving RDBMS with ML/DL algorithms. They have introduced multi-dimensional array-like indices for database tables, the improved query optimizer. It uses SimSQL which scales to large model size and can outperform TensorFlow. It shows that model parallelism for RDBMS based ML system can be scaled well and perform better than even Tensorflow.

The paper LARS\cite{lars} focuses on speeding up deep neural network training. Their approach focuses on increasing batch size using Layer-wise Adaptive Rate Scaling(LARS) for efficient use of massive resources. They train Alexnet and Resnet-50 using the Imagenet-1k dataset while maintaining its state of the art accuracy. However, they successfully increase the batch size to larger than 16k, thereby, reducing the training time of 100 epoch Alexnet from hours to 11 minutes and the 90-epoch ResNet-50 from hours to 20 minutes. Their system, for 90 epoch Resnet 50 returns an accuracy of 75.4\% with a batch size of 32k but reduces to 73.2\% when it is 64k. Their system clearly shows the extent to which large scale computers are capable of accelerating the training of a DNN by using massive resources with standard Imagenet-1k.

In LAMB\cite{lamb} they use a novel layerwise adaptive large batch optimization technique called LAMB. While LARS works well for Resnet, it performs poorly for attention models like BERT\cite{bert}. However, empirical results showed superior results with LAMB across systems like BERT with very little hyperparameter tuning. BERT could be trained with huge batch sizes of 32868 without degradation in performance. Hence, the training time of BERT reduced from 3 days to 76 minutes.

%% file: methodology.tex
\section{Methodology}
In this section, we define the architecture of our convolutional neural network at first. After that, we discuss LARS\cite{lars} optimization technique in details.

\subsection{Model Architecture}
\begin{figure}[h!]
  \includegraphics[width=\columnwidth]{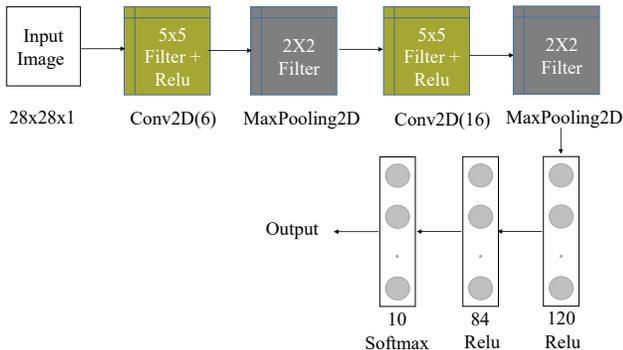}
  \caption{Architecture of our Convolutional Neural Network}
  \label{fig:cnn-1}
\end{figure}

Our CNN model consists of 2 convolution layers followed by 3 fully connected
layers including the classification layer. The first convolution layer has 6 filters of size 5 × 5 with
zero padding. The second convolution layer has 16 filters of size 5×5 with zero padding. There are two MAX Pooling layers with 2x2 filter. Each pooling layer follows one convolution layer. The first and second fully connected layers have dimensions 120 and 84 respectively, while the third fully connected
layer is of size 10. We apply ReLU activation function for all layers except the last layer, which has softmax activation. We apply cross-entropy loss function in the last layer. We don\textquotesingle t apply any dropout to any layer. The architectural structure of our convolutional neural network is shown in Figure \ref{fig:cnn-1}.

\subsection{LARS Optimizer}
Standard Stochastic Gradient Descent (SGD)\cite{sgd} uses same Learning Rate (LR) in each layer. The problems are two-fold. Firstly, if the learning rate is small, the convergence of training might take too much time. Sometimes, it might not even converge. Secondly, if the learning rate is high, the training process might end up with under-fitting of the model. In order to prevent under-fitting, the initial phase of training should be highly sensitive to the weight initialization and initial learning rate\cite{lars}. LARS\cite{lars} claims that the L2-norm of weights and gradients varies significantly between weights and biases, and between different layers. Training might become unstable if learning rate is large compared to this ratio in some layers. In order to solve this problem, there is a popular approach called \textit{learning rate warm-up}. This approach starts with a small learning rate which can be used for any layer and slowly increases the learning rate.

Instead of maintaining a single global learning rate, LARS maintains separate local learning rate for each layer. Equation \ref{eq:1} is used to calculate the weight using the local and global learning rate.
\[ \Delta w_{t}^{l} = \gamma * \lambda^l * \nabla{L(w_t)} \label{eq:1} \tag{1} \]
In Equation \ref{eq:1}, $\gamma$ is the global learning rate and $\lambda^l$ is the layer-wise local learning rate. Local learning rate for each layer is defined through a trust co-efficient which is much smaller than 1. Trust co-efficient defines how much a layer can be trusted to change its weight during update. Local learning rate can be calculated using Equation \ref{eq:2}.
\[ \lambda^{l} = \eta * \frac{||w^l||}{||\nabla{L(w^l)}||} \label{eq:2} \tag{2} \]
In Equation \ref{eq:2}, $\eta$ is the trust coefficient. In case of Stochastic Gradient Descent, a term known as \textit{weight decay}\cite{weight-decay} can be used. Equation for calculating local learning rate can also be extended to balance the learning rate with weight decay according to Equation \ref{eq:3}.
\[ \lambda^{l} = \eta * \frac{||w^l||}{||\nabla{L(w^l)}|| + \beta * ||w^l||} \label{eq:3} \tag{3} \]
In Equation \ref{eq:3}, $\beta$ is the weight decay term. Local learning rate helps to partially eliminate vanishing and exploding gradient problems\cite{lars}.

%% file: evaluation.tex
\section{Experimental Evaluation}
In this section, we define evaluation strategy and report experimental results. At first, we show the configuration of our machine. After that, we state evaluation metric, evaluation dataset and parameter settings. Finally, we graphically report experimental results.

\subsection{System Configuration}
We run our experiments in a Intel(R) Core(TM) i7-4790 CPU. The CPU frequency is 3.6 GHz and has 8 cores. L1d, L1i, L2, and L3 caches are 32K, 32K, 256K and 8192K respectively. Operating system version is Ubuntu 18.04.3 LTS x86\_64. We use 4 parallel batches to investigate the performance in a parallel and distributed setting.
\subsection{Evaluation Metric and Parameter Setting}
{\bf \em Dataset Used:} MNIST Dataset\cite{mnist-data}

{\bf \em Evaluation Metrics:}

We used three different evaluation metrics to evaluate the performance of our system which include Test Accuracy, Train Accuracy, and Generalization Error.

\textbf{Generalization Error} indicates the difference between training accuracy and test accuracy. Generalization error is high when a model gains very high training accuracy but low test accuracy. Normally, increasing batch size in stochastic gradient descent results in large generalization error. We select this evaluation metric because the goal of our system is to reduce the generalization error.

{\bf \em Parameter Setting:} The values of various parameters used for our CNN model training is reported in table \ref{table:param-values}.

\begin{table}[hbt!]
\begin{center}
\begin{scriptsize}
\begin{tabu} to \columnwidth { | c | c | }
\hline
Parameters & Values \\
\hline
Initial Learning rate & 0.01 \\
Learning rate Decay & 0.0001 \\
Weight Decay & 0.0001 \\
Momentum & 0.9 \\
Trust Coefficient & 0.001 \\
Number of Parallel Batches & 4 \\
\hline
\end{tabu}
\end{scriptsize}
\end{center}
\caption{Values of various parameters used during model training}
\label{table:param-values}
\end{table}

\subsection{Experimental Results}
\begin{figure}[h!]
  \includegraphics[width=\columnwidth]{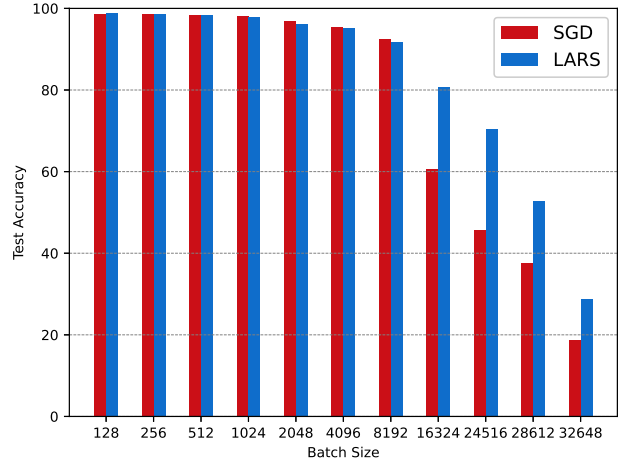}
  \caption{Test Accuracy of SGD and LARS Optimizers with Increasing Batch Size}
  \label{fig:acc-test}
\end{figure}

This subsection reports all experimental results in a graphical way. Figure \ref{fig:acc-test} compares the test accuracy achieved by both SGD\cite{sgd} and LARS\cite{lars} optimizers. From the figure, it is clear that both approaches perform extremely good for small batch sizes. Both optimizers maintain a test accuracy of around 99\% until batch size reaches 1024. After that, test accuracy starts decreasing gradually although they maintain an accuracy of more than 90\% until batch size reaches 8000. The difference between performances of these two optimizers is observable after that. The test accuracy of SGD optimizer starts dropping significantly once batch size reaches 16000 while LARS optimizer maintains much higher test accuracy compared to SGD. Test accuracy for SGD drops below 40\% after 28000 batch size while LARS optimizer maintains more than 55\% test accuracy for similar batch size. Test accuracy for both optimizers drops significantly once batch size reaches 32000 although LARS optimizer performs better yet.

\begin{figure}[h!]
  \includegraphics[width=\columnwidth]{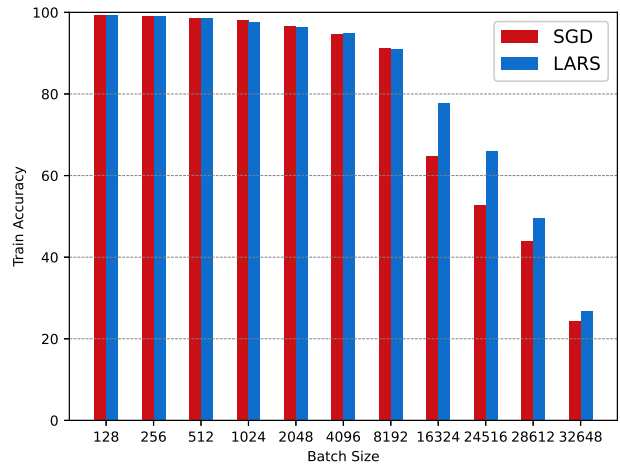}
  \caption{Training Accuracy of SGD and LARS Optimizers with Increasing Batch Size}
  \label{fig:acc-train}
\end{figure}

Figure \ref{fig:acc-train} compares the training accuracy achieved by both SGD and LARS. Similar to test accuracy in Figure \ref{fig:acc-test}, train accuracy is also higher for both approaches in case of small batch sizes. It is noticeable that train accuracy for SGD is better than its test accuracy. Still, it cannot outperform LARS for large batch sizes. Similar to test accuracy, train accuracy for SGD also decreases significantly once batch size reaches 16000. LARS outperforms SGD for both train accuracy and test accuracy if batch size is larger than or equal to 16000.

\begin{figure}[h!]
  \includegraphics[width=\columnwidth]{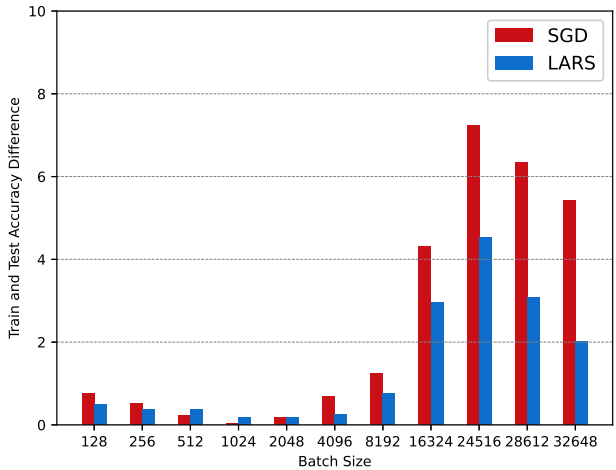}
  \caption{Generalization Error of SGD and LARS Optimizers with Increasing Batch Size}
  \label{fig:generalize-error}
\end{figure}

Figure \ref{fig:acc-train} compares the generalization error for both SGD and LARS optimizers. As we stated earlier, generalization error is very important in our evaluation because it indicates whether a model is consistent in train accuracy and test accuracy. More generalization error indicates more inconsistency. That\textquotesingle s why lower generalization error is always desired. Figure \ref{fig:generalize-error} indicates that generalization error for SGD is much higher than LARS optimizer. Although generalization error for SGD is within a desired range for batch sizes smaller than 8000, it increases significantly after that. It can be stated that generalization error for LARS is much smaller than SGD for batch sizes larger than 8000.

%% file: conclusion.tex
\section{Conclusion}
The encouragement towards training deep learning models with growing amount of data has been discussed thoroughly. Processing huge amount of data leads to so many real-time issues. The inability to use large batch size to utilize full system resources in highly configured environments is one of those issues which we focus here. We compare test accuracy of SGD and LARS optimizers for various batch sizes in a distributed deep learning setting using SystemML. For the ease of our evaluation, we choose a simple convolutional neural network model and MNIST\cite{mnist-data} dataset for our experimental evaluation. Based on our evaluation, we can conclude that LARS optimizer can be used to train with large batch sizes with distributed machine learning systems such as SystemML.

%% file: future-work-challenges.tex
\section{Future Work and Challenges}
We plan to extend this task by performing similar experiments with a large deep learning model such ResNet\cite{resnet}. So far, we used MNIST\cite{mnist-data} dataset for our experiments. We are also interested in evaluating our approach with a larger dataset such as ImageNet dataset\cite{imagenet-data}. Since LAMB\cite{lamb} optimizer proved to perform excellent for even large models such as BERT\cite{bert}, our another goal is to evaluate the performance of LAMB optimizer with distributed machine learning framework, SystemML.

We faced many challenges when finishing this task. Most of the challenges got raised due to many bugs of our adopted machine learning framework, SystemML. The latest release of this framework has several issues and raised errors once we tried to utilize some of its features. Despite all of these issues with SystemML, we hope to utilize this framework for our future experiments with complex model and large dataset.

%% file: main.bbl
\begin{thebibliography}{10}\itemsep=-1pt

\bibitem{systemml}
Matthias Boehm, Michael~W. Dusenberry, Deron Eriksson, Alexandre~V.
  Evfimievski, Faraz~Makari Manshadi, Niketan Pansare, Berthold Reinwald,
  Frederick~R. Reiss, Prithviraj Sen, Arvind~C. Surve, and Shirish Tatikonda.
\newblock Systemml: Declarative machine learning on spark.
\newblock {\em Proc. VLDB Endow.}, 9(13):1425–1436, Sept. 2016.

\bibitem{imagenet-data}
J. Deng, W. Dong, R. Socher, L.-J. Li, K. Li, and L. Fei-Fei.
\newblock {ImageNet: A Large-Scale Hierarchical Image Database}.
\newblock In {\em CVPR09}, 2009.

\bibitem{bert}
Jacob Devlin, Ming-Wei Chang, Kenton Lee, and Kristina Toutanova.
\newblock Bert: Pre-training of deep bidirectional transformers for language
  understanding.
\newblock {\em ArXiv}, abs/1810.04805, 2019.

\bibitem{resnet}
Kaiming He, Xiangyu Zhang, Shaoqing Ren, and Jian Sun.
\newblock Deep residual learning for image recognition.
\newblock {\em 2016 IEEE Conference on Computer Vision and Pattern Recognition
  (CVPR)}, pages 770--778, 2016.

\bibitem{declarative-rdbms}
Dimitrije Jankov, Shangyu Luo, Binhang Yuan, Zhuhua Cai, Jia Zou, Chris
  Jermaine, and Zekai~J. Gao.
\newblock Declarative recursive computation on an rdbms: Or, why you should use
  a database for distributed machine learning.
\newblock {\em Proc. VLDB Endow.}, 12(7):822–835, Mar. 2019.

\bibitem{mnist-data}
Yann LeCun, Corinna Cortes, and CJ Burges.
\newblock Mnist handwritten digit database.
\newblock {\em ATT Labs [Online]. Available: http://yann.lecun.com/exdb/mnist},
  2, 2010.

\bibitem{mllib}
Xiangrui Meng, Joseph Bradley, Burak Yavuz, Evan Sparks, Shivaram Venkataraman,
  Davies Liu, Jeremy Freeman, DB Tsai, Manish Amde, Sean Owen, Doris Xin,
  Reynold Xin, Michael~J. Franklin, Reza Zadeh, Matei Zaharia, and Ameet
  Talwalkar.
\newblock Mllib: Machine learning in apache spark.
\newblock {\em J. Mach. Learn. Res.}, 17(1):1235–1241, Jan. 2016.

\bibitem{weight-decay}
Kensuke Nakamura and Byung{-}Woo Hong.
\newblock Adaptive weight decay for deep neural networks.
\newblock {\em CoRR}, abs/1907.08931, 2019.

\bibitem{sgd}
Sebastian Ruder.
\newblock An overview of gradient descent optimization algorithms.
\newblock {\em arXiv preprint arXiv:1609.04747}, 2016.

\bibitem{lamb}
Yang You, Jing Li, Jonathan Hseu, Xiaodan Song, James Demmel, and Cho{-}Jui
  Hsieh.
\newblock Reducing {BERT} pre-training time from 3 days to 76 minutes.
\newblock {\em CoRR}, abs/1904.00962, 2019.

\bibitem{lars}
Yang You, Zhao Zhang, Cho-Jui Hsieh, James Demmel, and Kurt Keutzer.
\newblock Imagenet training in minutes.
\newblock In {\em Proceedings of the 47th International Conference on Parallel
  Processing}, ICPP 2018, New York, NY, USA, 2018. Association for Computing
  Machinery.

\bibitem{apache-spark}
Matei Zaharia, Reynold~S. Xin, Patrick Wendell, Tathagata Das, Michael
  Armbrust, Ankur Dave, Xiangrui Meng, Josh Rosen, Shivaram Venkataraman,
  Michael~J. Franklin, Ali Ghodsi, Joseph Gonzalez, Scott Shenker, and Ion
  Stoica.
\newblock Apache spark: A unified engine for big data processing.
\newblock {\em Commun. ACM}, 59(11):56–65, Oct. 2016.

\end{thebibliography}
